\documentclass{article}

\usepackage{microtype}
\usepackage{graphicx}
\usepackage{subcaption}
\usepackage{booktabs} 

\usepackage{hyperref}

\usepackage[preprint]{icml2026}

\usepackage{amsmath}
\usepackage{amssymb}
\usepackage{mathtools}
\usepackage{amsthm}

\usepackage[capitalize,noabbrev]{cleveref}

\theoremstyle{plain}

\theoremstyle{definition}

\theoremstyle{remark}

\usepackage[textsize=tiny]{todonotes}

\usepackage{enumitem}

\defcitealias{kimik2}{Kimi, 2025}
\defcitealias{DeepseekR1}{DeepSeek, 2025}

\icmltitlerunning{General learned delegation by clones}

\begin{document}

\twocolumn[
  \icmltitle{General learned delegation by clones}



  \icmlsetsymbol{equal}{*}

  \begin{icmlauthorlist}
    \icmlauthor{Darren Li}{qzc,wechat}
    \icmlauthor{Meiqi Chen}{wechat}
    \icmlauthor{Chenze Shao}{wechat}
    \icmlauthor{Fandong Meng}{wechat}
    \icmlauthor{Jie Zhou}{wechat}
  \end{icmlauthorlist}

  \icmlaffiliation{qzc}{Qiuzhen College, Tsinghua University}
  \icmlaffiliation{wechat}{Pattern Recognition Center, WeChat AI, Tencent Inc}

  \icmlcorrespondingauthor{Fandong Meng}{fandongmeng@tencent.com}

  \icmlkeywords{Large language models, Reinforcement learning}

  \vskip 0.4in
]



\printAffiliationsAndNotice{}  

\begin{abstract}
  Frontier language models improve with additional test-time computation, but serial reasoning or uncoordinated parallel sampling can be compute-inefficient under fixed inference budgets. We propose \textbf{SELFCEST}, which equips a base model with the ability to spawn same-weight clones in separate parallel contexts by agentic reinforcement learning.
Training is end-to-end under a global task reward with shared-parameter rollouts, yielding a learned controller that allocates both generation and context budget across branches.
Across challenging math reasoning benchmarks and long-context multi-hop QA, SELFCEST improves the accuracy-cost Pareto frontier relative to monolithic baselines at matched inference budget, and exhibits out-of-distribution generalization in both domains.
\end{abstract}

\section{Introduction}

Test-time scaling has emerged as a primary driver for capability improvements in large language models (LLMs) \cite{snell2024scalingllmtesttimecompute}.
Techniques such as self-consistency \cite{wang2023selfconsistency}, explicit branching search over chain of thought (CoT) reasoning \cite{yao2023treethoughts}, longer reasoning \citepalias{DeepseekR1}, and, more recently, agentic reasoning \citepalias{kimik2} can substantially improve accuracy over single-pass decoding. However, these improvements often come at significant efficiency costs under realistic inference budgets.
Serial CoT increases latency and token usage, while uncoordinated parallel methods spend compute by sampling many redundant reasoning paths and only aggregate at the end. These inefficiencies become sharper in long-context regimes, where effective usable context can be far smaller than the nominal context window and models struggle to robustly use information appearing mid-context \cite{liu2023lostmiddle,hsieh2024rulerwhatsrealcontext}.

\begin{figure}
    \centering
    \includegraphics[width=\linewidth]{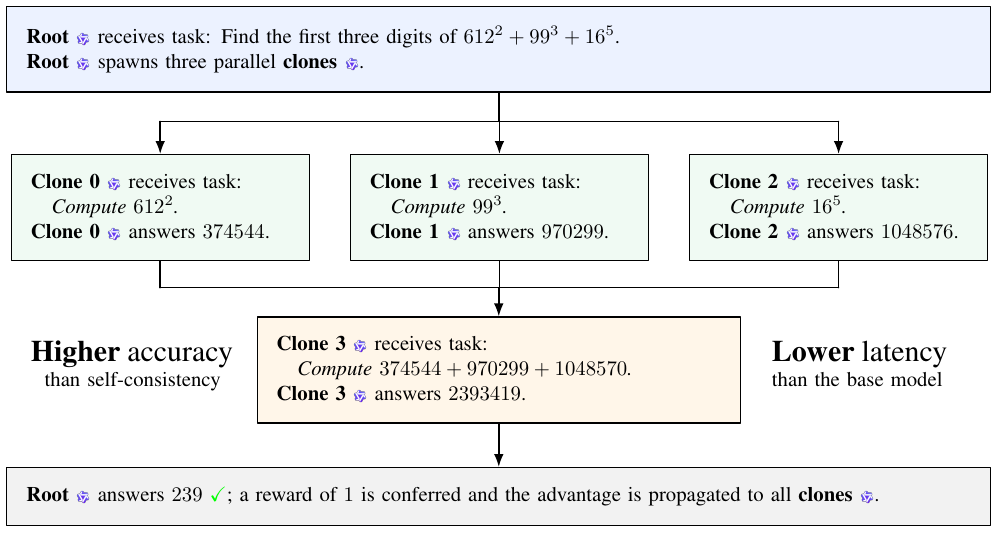}
\caption{A model trained with SELFCEST acts as a root agent, coordinateing helper clones (instantiated on the same parameters) for sub-tasks in parallel and then in series, successfully generalizing outside training data and outperforming various other methods in accuracy and token consumption.}
\vspace{-12pt}
    \label{fig:Fig1}
\end{figure}

Figure \ref{fig:Fig1} makes this distinction concrete in an arithmetic example. Prior parallel reasoning paradigms would approach this task by 
sampling multiple complete solution attempts or branching over alternative reasoning paths, each carrying the full context.
In contrast, the root agent in Figure \ref{fig:Fig1} explicitly spawns helper clones with narrowly scoped sub-tasks (e.g., computing individual terms), allocates separate context budgets to each clone, and aggregates their intermediate results to determine the next step. This example highlights a central bottleneck in test-time scaling: effective gains require not parallelism alone, but learned decisions about \emph{what} to parallelize and \emph{how} partial results should be recombined under a fixed inference budget.

A large body of prior work aims to tackle this inefficiency by making inference-time computation conditional rather than uniformly applied; a line of attack is hierarchical or multi-agent formulations by explicit decomposition -- splitting a problem into subproblems and coordinating specialized roles -- to improve exploration and credit assignment in long-horizon settings \cite{hu2025divideconquergroundingllms,ma2025coevolvingyoufinetuningllm}. This motivates studying whether this decomposition \emph{itself} can be treated as a first-class, learnable inference primitive for compute-aware reasoning. Adaptive Parallel Reasoning (APR) takes an important step in this direction by introducing spawn and join operations and training a controller for solving Countdown \cite{pan2025learningadaptiveparallelreasoning}. However, APR is evaluated in a specialized setting and does not establish policies that generalize across long-context reading and long-horizon reasoning. We seek to target \emph{general learned delegation}, emphasizing a distinct bottleneck: allocating limited context bandwidth and reasoning budget to the right sub-problems while keeping the root context compact.


We pursue this setting within the increasingly prominent tool-calling paradigm for LMs \cite{openai2023functioncalling,yao2022react,schick2023toolformer}, and propose \mbox{\textbf{SELFCEST}} (\textbf{SELF}-reinfor\textbf{CE}ment by \textbf{S}hared-weight \textbf{T}raining). 
SELFCEST introduces a single custom tool that allows a root agent to spawn same-weight helper clones in separate contexts.
As illustrated in Figure \ref{fig:Fig1}, the root agent learns to coordinate these clones by assigning sub-tasks, allocating context to each branch, and selectively joining returned intermediate states to form the final solution.
All agents \emph{share parameters} and are trained end-to-end with pure reinforcement learning under a global task reward, enabling the policy to learn what to compute in parallel as the root, and how to coordinate with the root and what information to send back as the clone. To address the challenge of credit assignment that arises in multi-agent reinforcement learning (MARL) at this scale \cite{NIPS2003_c8067ad1}, we devise and analyze reward shaping heuristics to determine a clone gradient update that yields stable training.

We validate SELFCEST on an array of long-reasoning and long-context tasks, measuring performance as a function of a fixed inference budget. On complex arithmetic benchmarks emphasizing task decomposition SELFCEST achieves a dramatic Pareto improvement over monolithic baselines and uncoordinated parallel inference, improving accuracy yet with lower end-to-end latency; across MATH-Hard \cite{hendrycks2021measuringmathematicalproblemsolving}, AIME, 2WikiMultiHopQA \cite{ho2020constructingmultihopqadataset}, and MuSiQuE \cite{trivedi2022musiquemultihopquestionssinglehop} improvements pointing to similar enhanced reasoning capabilities at matched total inference cost are also seen. 
Qualitative analyses indicate that the learned controller performs non-trivial allocation strategies, including selective evidence gathering in long contexts and adaptive branching on harder instances, suggesting that scale may unlock increasingly general delegation behaviors.

In summary, our contributions are: (i) a minimal tool-based formulation of learned delegation with shared-weight clones; (ii) an end-to-end RL training recipe for adaptive multi-thread inference under a unified cost budget; (iii) a credit-assignment analysis and stabilizing shaping variants for shared-parameter multi-agent rollouts; and (iv) empirical evidence of improved accuracy--cost trade-offs on long-context and long-reasoning tasks.

The rest of the paper proceeds as follows. We first formalize the global Markov decision process (MDP) to show that policy-gradient estimates remain unbiased when workers share parameters and discuss reward shaping in the context of credit assignment to enhance stability for less capable base models. We then present our experiments to demonstrate that the learned delegation policy generalizes out-of-domain and out-of-scale, suggesting that SELFCEST offers a practical post-training path toward compute-aware frontier models. 

\section{Related work}

\paragraph{Optimizing test-time scaling.} The inference-side efficiency of autoregressive decoding can be improved without changing the task-level policy, such as speculative decoding \cite{leviathan2023speculativedecoding}, draft and verify \cite{Zhang_2024}, and auxiliary decoding heads \cite{cai2024medusasimplellminference}. Transformer variants, e.g. early exit \cite{zhou2020bertlosespatiencefast} and depth-adaptive computation \cite{liu2020fasterdepthadaptivetransformers}, aim to halt computation once intermediate predictions are sufficiently confident, trading off accuracy for latency and FLOPs on easier inputs.
Related approaches prune tokens or selectively skip computations to reduce both depth-wise and width-wise cost \cite{he2021magicpyramidacceleratinginference}, and more recent conditional-computation schemes for language models route a limited subset of tokens through expensive layers to enforce a fixed compute budget while preserving quality \cite{raposo2024mixtureofdepthsdynamicallyallocatingcompute}.
These inference and architectural optimization methods are largely orthogonal to our setting: they reduce the cost of executing a given decoding policy, and can be composed with training-time approaches that learn \emph{what} to execute.

\paragraph{Agentic reinforcement learning.} 
Recent work has explored reinforcement learning for agentic language models that interact with environments, tools, or memory over multiple turns \cite{wang2023voyager, zhou2024archer}. 
These approaches demonstrate that language models can learn to decompose tasks, invoke tools, and revise intermediate states, but typically rely on hand-designed prompting strategies, fixed agent roles, or supervised traces \cite{shinn2023reflexion, schick2023toolformer}.
More recent RL-based approaches train agents end-to-end with environment feedback, enabling adaptive behavior across long horizons \cite{qi2024webrl, yuan2025reinforce}. 
However, most prior agentic RL work considers a single active agent interacting with tools sequentially, rather than coordinating multiple parallel reasoning threads under a shared budget.
Our work extends agentic RL to a multi-threaded inference setting, where the agent must learn not only which actions to invoke, but also when to spawn parallel workers, how much context to allocate to them, and how to aggregate their returned states to solve the task efficiently.

\begin{figure*}[t]
    \centering
    \includegraphics[width=\linewidth]{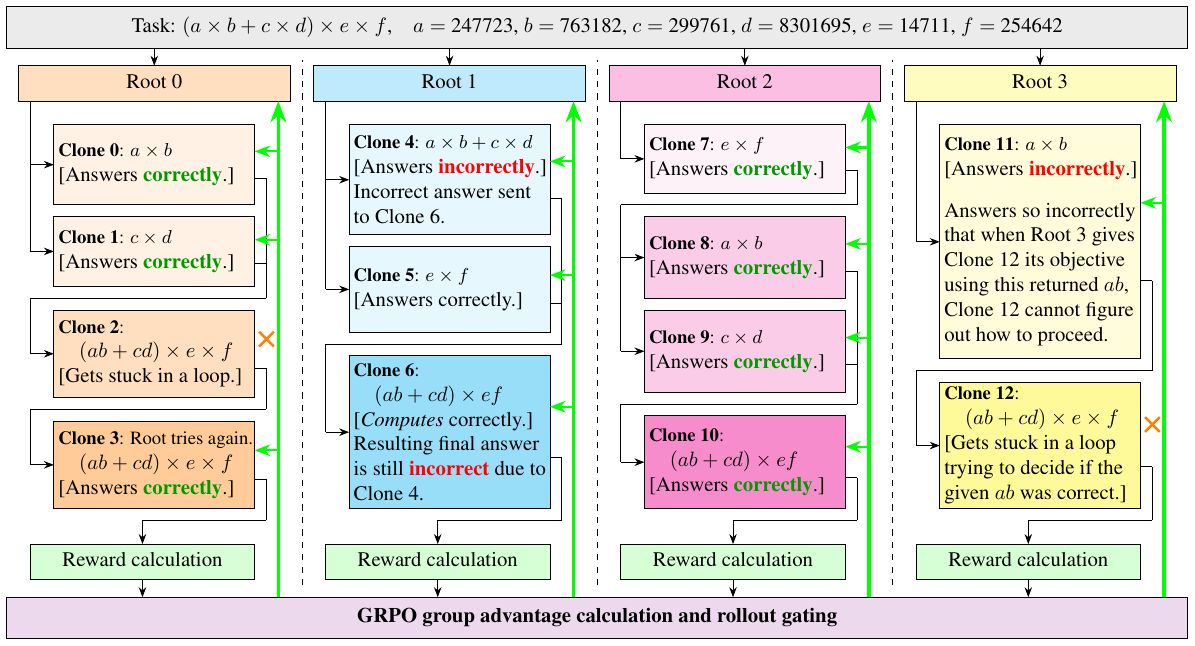}
\caption{Illustration of one training step of SELFCEST using GRPO. A positive advantage is conferred to roots 0 and 2 -- the roots returning correct final answers -- and clones 0, 1, 3, 7, 8, 9, and 10; a negative advantage is conferred to roots 1 and 3, as well as clones 4, 5, 6, and 11; clones 2 and 12 is conferred zero advantage and their trajectories do not participate in backpropagation. Despite the correct computation of Clone 5 ultimately getting a negative advantage, this crude gating is enough to guarantee stability.}
    \label{fig:Fig2}
\end{figure*}

\paragraph{Credit assignment problem.} Even with shared parameters, using a single team-level return to update the joint trajectory comprising of all clone rollouts leads to a multi-agent credit assignment problem: individual clones may receive spurious credit when the team succeeds -- i.e. the root's final answer is correct -- despite their failures. This issue is well-studied in cooperative MARL with global rewards \cite{NIPS2003_c8067ad1}; proposed mitigation techniques include isolating each agent's contribution by counterfactual baselines \cite{foerster2024counterfactualmultiagentpolicygradients} and explicit value factorization, e.g. value decomposition networks \cite{sunehag2017valuedecompositionnetworkscooperativemultiagent} and QMIX \cite{rashid2018qmixmonotonicvaluefunction}. Additionally, our setting exhibits delayed-reward temporal credit assignment across long token sequences, for which process rewards have been used to achieve step-by-step reinforcement, such as in MATH-Shepherd \cite{wang2024mathshepherdverifyreinforcellms}. 
Due to resource constraints in our setting applying these high-quality assignment techniques is difficult; we propose rollout gating as a lightweight heuristic, and leave integrating these methods to future work. 

\section{Method}

\paragraph{Theoretical setup.} If we model the entire root and clone process as one episodic MDP with a scheduler, then we can let the global state $s_t$ contain the root transcript so far, the set of live clone rollouts (their transcripts, status, which tool-call they came from), and a pointer to whose turn is active. At each step, the action is one token from the currently-active rollout. A ``spawn clone'' tool call is just a special token sequence that deterministically creates a new rollout; the scheduler switches control to that rollout until it returns. By having all clones share parameters $\theta$, the joint trajectory $\tau$ has log probability
\vspace{-2pt}
\[\log p_\theta(\tau)=\sum_{\text{rollouts }i}\ \sum_{t\in i}\ \log \pi_\theta\left(a_t^{(i)} \left| h_t^{(i)}\right.\right)\]
where $h_t^{(i)}$ is the conditioning for rollout $i$, i.e. the prompt from the root thread, generated tokens, and system prompts.

For an objective $J(\theta)=\mathbb E_\theta[R(\tau)]$, the policy gradient -- with importance sampling \cite{schulman2017trustregionpolicyoptimization}, clipping \cite{schulman2017proximalpolicyoptimizationalgorithms}, and KL divergence penalty \cite{NEURIPS2020_1f89885d} terms omitted for clarity -- is
\begin{align*}\nabla J(\theta)&=\mathbb E_\theta\left[ R(\tau)\nabla \log p_\theta(\tau)\right]\\
&=\mathbb E_\theta\left[\sum_iR(\tau)\sum_{t\in i} \nabla \log \pi_\theta\left(a_t^{(i)} \left| h_t^{(i)}\right.\right)\right],\end{align*}
and as the gradient due to the joint MDP is identical to the sum of gradients of each independent rollout (root plus all clones), the reward can be directly propagated to all clones, enabling true end-to-end training.

\paragraph{GRPO advantage.} We now need to obtain an advantage estimator $A(\tau)$ such that $A(\tau) \nabla \log p_\theta(\tau)$ is an unbiased estimator of $\nabla J(\theta)$ of lower variance. Our starting point, to circumvent the technical complexity of defining the critic model necessary for proximal policy optimization, is group relative policy optimization \cite{shao2024deepseekmathpushinglimitsmathematical}. The GRPO outcome supervision advantage is calculated from $G$ trajectories $\{\tau[1], \tau[2], \dots, \tau[G]\}$ with their corresponding rewards $\mathbf{r}=\{r_1,r_2,\dots,r_G\}$ as
\[A(\tau[i])=\frac{r_i-\text{mean}(\mathbf r)}{\text{std}(\mathbf r)}.\]

\paragraph{COMA-style counterfactual baselines.} The rest of this method is improving stability by a more fine-grained baseline. Specifically, if $b^{(i)}_t$ is the rollout $i$ baseline at time $t$ that can depend on everything but the current action $a^{(i)}_t$, then
\[\nabla J(\theta)=\mathbb E_\theta\left[\sum_{i, t\in i}\nabla \log \pi_\theta\left(a_t^{(i)} \left| h_t^{(i)}\right.\right) \left(R(\tau)-b_t^{(i)}\right)\right]\]
as the contribution to the expectation $\mathbb E_\theta[\pi_\theta(a|h)b(\hat{a},h)]$ vanishes. Ideally, we would define a centralized action-value for each token decision
\[Q_t^{(i)}\left(h_t^{(i)}, a_t^{(i)}\right) \stackrel{\text{def}}{=}
\mathbb E_\theta\left[ R(\tau)\middle| h_t^{(i)}, a_t^{(i)}\right]\]
to obtain the {counterfactual baseline} averaging over what agent $i$ \emph{would} have done at that same decision point
\[b_t^{(i)}\left(h_t^{(i)}\right)=\sum_{a'} \pi_\theta\left(a' \left| h_t^{(i)}\right.\right) Q_t^{(i)}\left(h_t^{(i)}, a'\right).\]
This yields a per-token, per-rollout counterfactual advantage $A^{(i)}_{t,\text{cf}}$. The resulting estimator for the actor update
\vspace{-2pt}
\[\nabla J(\theta)=\mathbb E_\theta\left[\sum_{i}\sum_{t\in i}\nabla \log \pi_\theta\left(a_t^{(i)} \left| h_t^{(i)}\right.\right) A^{(i)}_{t,\text{cf}}\right]\]
\vspace{-12pt}

is unbiased, but relies on approximating $Q^{(i)}_t$, a centralized critic over a variable set of rollouts and long token histories, and re-encounters the same practical obstructions that we sought to alleviate with GRPO.

\paragraph{Difference rewards.} We opt for baselines at the rollout granularity rather than token granularity. Let $\tau_{-i}$ denote the joint trajectory with clone $i$'s internal actions replaced by a ``null'' policy. If we can compute $R(\tau_{-i})$ exactly, then the difference reward
\[D_i(\tau) \stackrel{\text{def}}{=} R(\tau)- R(\tau_{-i})\]
is a per-rollout credit signal, giving the update
\[\nabla J(\theta)=\mathbb E_\theta\left[\sum_{i}D_i(\tau)\sum_{t\in i}\nabla \log \pi_\theta\left(a_t^{(i)} \left| h_t^{(i)}\right.\right) \right].\]

\paragraph{Rollout gating.} In practice, even $R(\tau_{-i})$ is expensive because it requires recomputing not only clone $i$ but all root actions afterwards, including all clones that the root spawned afterwards. We approximate $D_i(\tau)$ as a single gate on the GRPO advantage, applying a lightweight rollout gating rule that downweights clones that are very likely non-contributing due to protocol violations. Concretely, for each clone $i$, we compute a gate $w_i(\tau)\in [0,1]$ and use the update
\[\nabla J(\theta)=\mathbb E_\theta\left[\sum_{i}w_i(\tau)A(\tau)\sum_{t\in i}\nabla \log \pi_\theta\left(a_t^{(i)} \left| h_t^{(i)}\right.\right) \right].\]
Instantiations of $w_i(\tau)$ we tested are
\begin{itemize}
  \item \textbf{Hard protocol gate.}
\[w_i = {1}[\text{clone }i\text{ returns a parseable final answer}].\]
  \item \textbf{Soft quality gate.} $w_i=\sigma(\alpha\ \text{Score}_i)$, for a score with penalties based on indicators such as missing/empty return markers, or termination due to technical limits like the per-rollout token generation limit.
  \item \textbf{Use-based gate.} $w_i=1[\text{root used clone }i]$, where we detect ``use'' by a deterministic substring match in the root's final reasoning of the clone's return.
\end{itemize}
Despite the use-based gate approximating marginal contribution best, it performed badly due to misattributing credit when the root coincidentally reproduces the same content; our final method uses the hard protocol gate, in effect identical to protocol-based gradient masking.

\paragraph{Bias due to rollout gating.} We emphasize that rollout gating introduces bias relative to the unbiased global-return gradient. Rollout gating corresponds to optimizing a reweighted policy-gradient estimator; it reduces variance and suppresses pathological trajectories, but is biased with respect to the original global-return objective because $w_i$ depends on the sampled clone trajectory. Experimentally, the elimination of obviously non-contributing rollouts is critical for training stability in our testing. Similar bias-stability tradeoffs appear in difference-advantage style credit assignment methods \cite{pmlr-v162-li22w}.

\paragraph{Connection between counterfactual baselines and difference rewards.} Both difference rewards and COMA-style counterfactual baselines can be viewed as approximations to a common object: the expected return under a conditional counterfactual in which clone $i$'s decisions are marginalized out while all earlier context is held fixed. Let
\[b^{(i)}(\tau)=\mathbb E_{\tau'\sim \pi_\theta}\left[ R(\tau')\middle| \tau' \text{ agrees with }\tau\text{ before clone }i\right],\]
i.e. conditioned on $h^{(i)}$. Difference rewards approximate $\bar R_i(\tau)$ using a single deterministic counterfactual rollout $R(\tau_{-i})$, while COMA replaces this with an expectation over clone $i$'s actions under the current policy
\[\bar R_i(\tau)=\sum_{a'} \pi_\theta\left(a'\left| h^{(i)}\right.\right)\mathbb E_\theta\left[R(\tau') \left| h^{(i)}, a^{(i)}=a'\right.\right].\]
\vspace{-12pt}

Both yield unbiased gradients in the limit of exact counterfactual evaluation, but are costly to realize in our setting. Our rollout-gating heuristic can be interpreted as a coarse, low-variance approximation to this conditional counterfactual that suppresses gradients from evidently non-contributing clones.

\paragraph{Reward function.} Our reward function is global and is composed of a single accuracy term and two penalty terms
\begin{align*}R(\tau)=R_0(\tau)&-\text{Rt}_fL(\text{Rt}_\tau-\text{Rt}_{\text{th}},\text{Rt}_{\text{ramp}})\\&-\max_i\text{Cl}_fL(\text{Cl}_\tau^{(i)}-\text{Cl}_{\text{th}},\text{Cl}_{\text{ramp}})\end{align*}
where $R_0(\tau)$ is $1$ if the root answer is correct and $0$ otherwise, $\text{Rt}_{\text{th}},\text{Rt}_{\text{ramp}},\text{Rt}_{\text{f}}$ and $\text{Cl}_{\text{th}},\text{Cl}_{\text{ramp}},\text{Cl}_{\text{f}}$ are hyperparameters, and for a token excess $x$ and a ramp factor $r$ the penalty is $L(x,r)=\max\left(0,1-e^{-x/r}\right)$.

\section{Pure arithmetic experiments}

In this section, we introduce our custom arithmetic dataset, and describe our experimental results and ablations on this dataset.

\paragraph{Custom arithmetic dataset.} We construct a synthetic dataset inspired by MATH 401 \cite{yuan2023largelanguagemodelsperform}, consisting of a narrower scope but significantly harder problems. We omit all arithmetic operations harder than division like trigonometric functions and all floating point computation, because their computation represents a higher intrinsic difficulty. At the same time we extend the magnitude of the integers within the expressions to $10^{7}$ and the total number of arithmetic operations to $10$, and allow intermediate and final answers of magnitude up to $10^{16}$. For reproducibility, we release the script used to generate this dataset.

\paragraph{Experimental setup and hyperparameters.} The base model we do agentic RL on is Qwen3-4B-Instruct-2507, a SOTA dense non-thinking model in its weight class with strong tool-calling capabilities.

To implement the SELFCEST rollout collection and gated gradient calculation, we modify the tool agent loop of Volcano Engine Reinforcement Learning (VERL), with rollouting done using vLLM \cite{sheng2024hybridflow, kwon2023efficient}. Our hyperparameters are listed in Table \ref{table:hyperparameters}; we hold them constant for all later long reasoning/context experiments.

\begin{table}[]
  \caption{Hyperparameters.}
  \label{table:hyperparameters}
  \begin{center}
        \begin{tabular}{ll}
          \toprule
          Parameter name  & Value \\
          \midrule
          Optimizer & AdamW \\
          Learning rate & $\text{6}\cdot \text{10}^{\text{-7}}$ \\
          Warmup steps & 10 \\
          Groups per batch & 32 \\
          Tensor parallel & 8 \\
          \midrule
          Prompt length limit & 1024 \\
          Token generation limit & 1024 \\
          \midrule 
          GRPO group sampling & 4 \\
          Trajectories per batch & 128 \\
          Minibatch size & 8 \\
          GRPO clip ratio & 0.1 \\
          KL divergence penalty & $\text{5}\cdot \text{10}^{\text{-4}}$ \\
          Loss aggregation & Token mean \\
          \midrule
          Maximum tool use turns & 10 \\
          Maximum summary length & 256 bytes \\
          JSON repair penalty & 0.05 \\
          \midrule
          Root token penalty $\text{Rt}_{\text{th}},\text{Rt}_{\text{ramp}},\text{Rt}_{\text{f}}$ & 512, 256, 0.3 \\
          Clone token penalty  $\text{Cl}_{\text{th}},\text{Cl}_{\text{ramp}},\text{Cl}_{\text{f}}$ & 512, 512, 0.2 \\
          \bottomrule
        \end{tabular}
  \end{center}
  \vskip -0.1in
\end{table}

\paragraph{JSON repair penalty.} Due to minor formatting problems in the generated tool calling JSON, we introduce automated JSON repair to the tool call parser in VERL and apply a global penalty for malformed tool calls.

\begin{figure}[H]
    \centering
    \includegraphics[width=\linewidth]{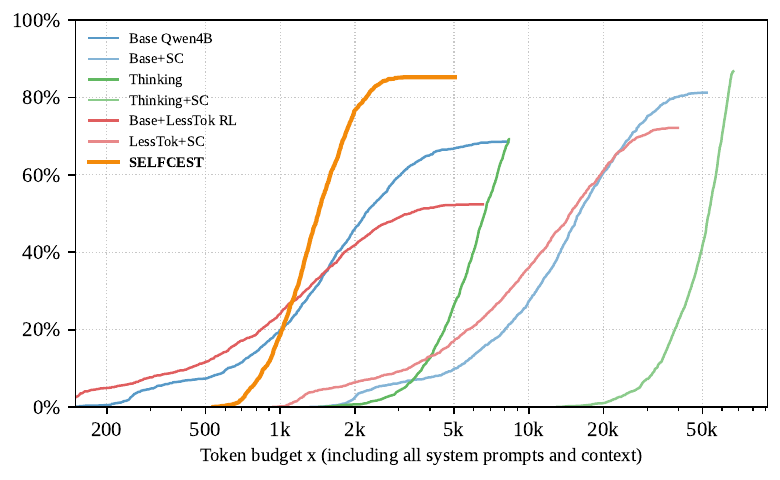}
\caption{SELFCEST solves more problems in less tokens.}
    \label{fig:Fig3}
\end{figure}

\paragraph{Results.} We compare our results against six baselines: three models -- the base Qwen3-4B-Instruct-2507, the thinking variant Qwen3-4B-Thinking-2507, and a fine tune of Qwen3-4B-Instruct-2507 on this exact dataset using the same reward function as SELFCEST but without the clone tool to see if lower token usage without costing accuracy could be obtained by pure reinforcement learning -- and a self-consistency variant obtained by parallel sampling x8 of all three of these models, as shown in Figure \ref{fig:Fig3}.

We remark that the counterintuitively worse performance above a \~{}1500 token budget of the fine-tune obtained by pure reinforcement learning is likely due to the aggressive root token penalty. However, due to resource constraints, we were unable to test at higher token generation limits; at the same token generation limit, decreasing the root token penalty makes performance \emph{worse} due to the increased noise arising from more rollouts terminated by the token generation limit.

A significantly more physical advantage is latency. 
We approximate latency by total \emph{generated} tokens (divided by 8 for self-consistency; in reality the self-consistency latency would be slightly higher as it is often decided by the longest rollouts).

\begin{table}[H]
  \caption{Average generated tokens.}
  \label{table:generatedtokens}
  \begin{center}
        \begin{tabular}{ll}
          \toprule
          Model & Average generated tokens \\
          \midrule
          Base Qwen3-4B & 1,924 \\
          Qwen3-4B-Thinking & 6,220 \\
          Qwen3-4B low token RL & 1,455 \\
          \midrule
          \textbf{SELFCEST} & \textbf{930} \\
          \bottomrule
        \end{tabular}
  \end{center}
  \vskip -0.1in
\end{table}

\begin{figure}[t]
    \centering
    \includegraphics[width=\linewidth]{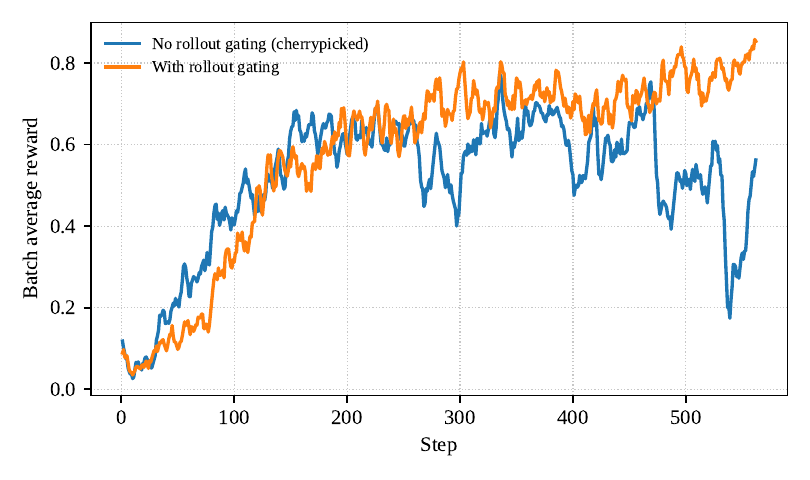}
\caption{Comparison of reward between our final training run and the run without rollout gating with the highest final reward.}
    \label{fig:Fig4}
\end{figure}

\paragraph{Significance of rollout gating.} We observe experimentally that rollout gating is indispensable for both training stability and final performance. To gauge the importance of rollout gating, we conducted 8 training runs without rollout gating on the same hyperparameters and otherwise same reward function. In 5 such runs the performance stayed worse than before training started, indicating a complete failure to learn; we observed in such runs a catastrophic forgetting of the tool call capabilities present in the base model. The other 3 runs demonstrate severe instability, with sudden reward drops. We present a graph of the mean batch reward over training steps observed in one such run in Figure \ref{fig:Fig4}.

\paragraph{Other ablations.} We provide a brief overview of the best performance observed over 4 independent training runs on the same hyperparameters for the different rollout gates in Section 3. For the soft quality gate, we used a baseline score of $5$ and deducted $3$ points for each of a missing return marker, an unfinished rollout, and a truncated answer returned to the root, for a worst case score of $-4$.

\begin{table}[H]
  \caption{Different rollout gates.}
  \label{table:generatedtokens}
  \begin{center}
        \begin{tabular}{ll}
          \toprule
          Rollout gate & Best final accuracy @4 \\
          \midrule
          \textbf{Hard protocol gate} & \textbf{0.86} \\
          Soft quality gate & 0.73 \\
          Use-based gate & 0.40 \\
          \bottomrule
        \end{tabular}
  \end{center}
  \vskip -0.1in
\end{table}

\subsection{Case studies of notable trajectories encountered}

\paragraph{Root correct despite multiple bad/gated clones.} We observed in 2 training runs a peculiar class of behavior in the early stages of training, specifically when the accuracy had not yet increased to the point where the longest clone penalty mattered: redundancy of clones by spawning as many as possible, all with different objectives.

In one instance of the form $(a+b)\times c+d-e+f$ and exact values $a=3483838$, $b=239$,  $c=5709526$,  $d=8803$, $e=5446472$,  $f=5530030$, the root spawned \emph{five} clones in parallel, one for every prefix:
\begin{itemize}[nosep]
    \item $a+b$;
    \item $(a+b)\times c$;
    \item $(a+b)\times c+d$;
    \item $(a+b)\times c+d-e$;
    \item $(a+b)\times c+d-e+f$.
\end{itemize}
The latter two clones failed due to the token generation limit and were rollout gated. The root gracefully recovered from this, detecting that the answer provided by the last two clones -- implemented as a truncation of its entire output for missing answer markers -- could not be used, and spawning a single clone using the intermediate answer of Clone 3 to successfully complete the calculation.

This behavior vanishes as training progressed, but the immediate mitigation of per-clone instability by the root is noteworthy.

\paragraph{Root incorrect due to a single bad aggregation step.} On the opposite end of the spectrum, we have instances where the root fails due to one single error, as hinted in in Figure \ref{fig:Fig2}. In this particularly bad instance of the form \[a\times b\times(c - d) + e\times f\times g + h\times i\times (j-k),\] 
the root successfully orchestrated 8 clones to calculate the three main terms $ab(c-d)$, $efg$, and $hi(j-k)$, but the last clone failed to calculate the correct sum.

\begin{table*}[]
  \caption{Comparison of SELFCEST against base models on datasets used in training. Columns correspond to accuracy ($\uparrow$); parenthesized values correspond to average token budget, including system prompts and context ($\downarrow$). Average token budget for Qwen3-4B-Thinking is unavailable due to a failure to replicate the reported performance on our available token generation length.}
  \label{table:mixedtraining}
  \begin{center}
        \begin{tabular}{llll}
          \toprule
        Model & Arithmetic & AIME & MATH-Hard \\
          \midrule
          Base Qwen3-4B & 0.68 (2028) & 0.47 (3928) & 0.50 (1124) \\
          Qwen3-4B-Thinking & 0.69 (6306) & \textbf{0.81} (N/A) & \textbf{0.66} (4913) \\
          \midrule
          \textbf{SELFCEST} & \textbf{0.83 (1705)} & 0.47 (\textbf{1850}) & 0.57 (\textbf{932}) \\
          \bottomrule
        \end{tabular}
  \end{center}
  \vskip -0.1in
\end{table*}

Despite these kinds of scenarios introducing a negative advantage to \emph{all} clones, not just the last one, it did not appear to have a significant effect on the policy. This stands in contrast to incoherent rollouts without rollout gating, which we observed to immediately crater the accuracy metric in the next training step. We observed that almost all clones in the 4 trajectories in the same GRPO group -- the other 3 of which succeeded -- had almost the same clone objectives in slightly different order. It is likely that even without strong process guidance, the negative gradient on the clones in the erroneous trajectory was partially absorbed by the positive gradient on similar clones in correct trajectories.

\section{Long reasoning, long context, and hybrid experiments}

In this section we present experimental results on an array of pre-existing datasets. This section is not as complete as we would like; more datasets, comparisons to the performance of other models, and ablations will be added soon.

\paragraph{Mixed math training.} To test the applicability across more varied forms of mathematics reasoning, we trained on a combination of the arithmetic dataset described in Section 4, the LightEval MATH-Hard dataset -- the subset of the MATH dataset \cite{hendrycks2021measuringmathematicalproblemsolving} with the Level 5 labeled difficult -- double weighted, and the AIME 1983-2024 dataset \cite{aime_1983_2024} triple weighted. Table \ref{table:mixedtraining} compares the validation accuracy of the resulting policy against base models.

\paragraph{Isolated AIME dataset training.} We observe that training only on the AIME dataset exhibits worse final performance, even if we train for significantly more epochs (7 instead of the effective 3 for the hybrid dataset.)

\begin{table}[H]
\caption{Comparison of resulting AIME performance against training dataset/epoch combinations.}
\vspace{-6pt}
\begin{center}
    \begin{tabular}{ll}
      \toprule
      Training dataset & AIME ($\uparrow$) \\
      \midrule
      Pure AIME, $\times 1$ & 0.30 \\
      Pure AIME, $\times 7$ & 0.39 \\
      Hybrid dataset & \textbf{0.47} \\
      \bottomrule
    \end{tabular}
\end{center}
\vskip -0.1in
\end{table}

This indicates that the synthetic arithmetic dataset is conducive to learning an effective delegation strategy as the root agent; SELFCEST matches the reported AIME performance of the base Instruct model in an average token budget of 1850 tokens, down from 3928 tokens by the base Instruct model.
We note that this is nowhere close to the \textbf{0.81} accuracy of the Thinking model. Our limited testing of the Thinking model at a 8192 token generation limit observes \textbf{0.20} accuracy and consumes an average 7987 tokens, i.e. the token budget necessary for the Thinking model to exhibit its reasoning capabilities in more difficult math problems far exceeds our compute resources. 

\paragraph{Multi-hop long context.} We validate SELFCEST on the 2WikiMultiHopQA \cite{ho2020constructingmultihopqadataset} and MuSiQuE \cite{trivedi2022musiquemultihopquestionssinglehop} multi-hop question answering benchmarks. Due to the limited context of our setup, we \emph{force} the model to spawn clones with a key that is expanded into the corresponding context entry to retrieve the necessary information; this is in effect an agentic retrieval-augmented generation pipeline \cite{singh2025agenticretrievalaugmentedgenerationsurvey}, trained end-to-end by reinforcement learning. We compare against the base Instruct model, given either the full context or an information retrieval tool.

\begin{table}[H]
\caption{Comparison of resulting SELFCEST against base Instruct model and an agentic RAG pipeline on multi-hop QA.}
  \begin{center}
        \begin{tabular}{lll}
          \toprule
        Model & 2WikiMultiHopQA & MuSiQue \\
          \midrule
          Base Qwen3-4B & 0.65 (1214) & 0.35 (2417) \\
          Base + tools & 0.23 (726) & 0.14 (860) \\
          \midrule
          \textbf{SELFCEST} & \textbf{0.70 (644)} & \textbf{0.51 (810)} \\
          \bottomrule
        \end{tabular}
  \end{center}
\end{table}

\section{Future work}

The natural next step is to improve credit assignment beyond heuristic rollout gating. As discussed in our analysis of clone credit assignment, more accurate estimators of the conditional counterfactual return $\bar R_i(\tau)$ could reduce both bias and variance in clone updates by better approximating a clone's marginal contribution given the joint trajectory. One promising direction is to replace gating with a learned contribution estimator. Recent work \cite{zheng2023judging} suggests that LLMs can serve as reliable judges or preference models for evaluating and comparing LLM outputs, and in our setting a judge could condition on the all root and clone rollouts to estimate whether -- and to what extent -- each clone influenced the final answer, yielding a soft approximation to $\bar R_i(\tau)$ without explicit counterfactual rollouts. Viewed another way, such a judge amortizes counterfactual evaluation into a trajectory-conditioned value function.

Separately, our current training recipe uses lightweight approximations to multi-agent credit assignment and budget control. Incorporating stronger MARL baselines (e.g., counterfactual or value-factorized methods) may further improve stability and sample efficiency, while tighter budgeted scheduling policies could explicitly optimize trade-offs among latency, token cost, and branching depth (e.g., when to spawn, how many clones to run in parallel, and how aggressively to truncate or summarize returned state). We leave these integrations to future work.

\section{Conclusion}

Test-time scaling is increasingly limited by the lack of learned structure in how that compute is spent, especially in long-context settings where bandwidth and attention over the prompt are scarce. This paper framed this bottleneck as a control problem: under a fixed inference budget, the model should learn what to parallelize, how much context to allocate, and how to recombine intermediate results to maximize task success.

We introduced \textbf{SELFCEST}, a minimal tool-based formulation of learned delegation in which a root agent can spawn same-weight helper clones in separate contexts. Training is end-to-end with reinforcement learning under a global task reward, turning multi-threaded inference into a learnable policy rather than a hand-designed scaffold. A central practical obstacle is credit assignment across shared-parameter rollouts; we therefore analyzed the unbiased global-return gradient in the joint MDP and showed how lightweight \textbf{rollout gating}, despite introducing bias, stabilizes training by suppressing evidently non-contributing trajectories. 

Empirically, SELFCEST improves the \textbf{accuracy-cost Pareto frontier} on challenging arithmetic, broader math reasoning, and long-context multi-hop QA by learning non-trivial allocation behaviors. Beyond the specific tool interface, our results suggest a broader direction: delegation as a first-class inference primitive for compute-aware reasoning. Rather than viewing parallelism as ``more samples,'' SELFCEST points toward policies that treat context bandwidth and generation as resources to be dynamically budgeted across specialized subcalls, and approaches the learned counterpart to classical decomposition strategies from \emph{reward} rather than prescription by fixed roles.

In summary, SELFCEST offers a simple and practical route to learned multi-thread inference: a post-training method that improves efficiency by teaching models the optimal way to allocate computation through delegation.

\section*{Impact Statement}

SELFCEST advances how language models allocate test-time computation, enabling more efficient multi-threaded inference under fixed budgets. The most direct positive impact is improved accessibility: if delegation policies can reach a target accuracy with fewer tokens/latency, they lower the cost of deploying capable reasoning and long-context QA on resource-constrained hardware (e.g., edge devices or small-scale services), reducing energy use per query and broadening who can run strong models. At the same time, making models better at decomposing tasks and coordinating parallel subcalls can amplify their effectiveness in downstream agentic settings, which may increase the scale and speed at which automated systems can operate; this raises familiar concerns around over-reliance, automation of low-quality analysis at higher throughput, and misuse in domains where stronger planning/search is harmful. Our work does not introduce new tools beyond model self-calls, but it can increase capability-per-compute. We view safety evaluation of delegated multi-agent behaviors (e.g., robustness, calibration, and misuse-oriented stress tests) and monitoring of cost--capability scaling as important complements to future deployments.

\bibliography{main}

@misc{kimik2,
      title={Kimi K2: Open Agentic Intelligence}, 
      author={{Kimi Team}},      year={2025},
      eprint={2507.20534},
      archivePrefix={arXiv},
      primaryClass={cs.LG},
      url={https://arxiv.org/abs/2507.20534}, 
      note          = {Full author list: see arXiv:2507.20534},
}

@misc{snell2024scalingllmtesttimecompute,
      title={Scaling LLM Test-Time Compute Optimally can be More Effective than Scaling Model Parameters}, 
      author={Charlie Snell and Jaehoon Lee and Kelvin Xu and Aviral Kumar},
      year={2024},
      eprint={2408.03314},
      archivePrefix={arXiv},
      primaryClass={cs.LG},
      url={https://arxiv.org/abs/2408.03314}, 
}

@article{DeepseekR1,
   title={DeepSeek-R1 incentivizes reasoning in LLMs through reinforcement learning},
   volume={645},
   ISSN={1476-4687},
   url={http://dx.doi.org/10.1038/s41586-025-09422-z},
   DOI={10.1038/s41586-025-09422-z},
   number={8081},
   journal={Nature},
   publisher={Springer Science and Business Media LLC},
   author={{Deepseek-AI Team}},
   year={2025},
   month=sep, pages={633–638},
   note          = {Full author list: see arXiv:2501.12948},
}

@misc{openai2023functioncalling,
  author = {{OpenAI}},
  title = {Function calling and other API updates},
  year = {2023},
  month = {6},
  url = {https://openai.com/index/function-calling-and-other-api-updates}
}

@misc{hu2025divideconquergroundingllms,
      title={Divide and Conquer: Grounding LLMs as Efficient Decision-Making Agents via Offline Hierarchical Reinforcement Learning}, 
      author={Zican Hu and Wei Liu and Xiaoye Qu and Xiangyu Yue and Chunlin Chen and Zhi Wang and Yu Cheng},
      year={2025},
      eprint={2505.19761},
      archivePrefix={arXiv},
      primaryClass={cs.AI},
      url={https://arxiv.org/abs/2505.19761}, 
}

@misc{pan2025learningadaptiveparallelreasoning,
      title={Learning Adaptive Parallel Reasoning with Language Models}, 
      author={Jiayi Pan and Xiuyu Li and Long Lian and Charlie Snell and Yifei Zhou and Adam Yala and Trevor Darrell and Kurt Keutzer and Alane Suhr},
      year={2025},
      eprint={2504.15466},
      archivePrefix={arXiv},
      primaryClass={cs.AI},
      url={https://arxiv.org/abs/2504.15466}, 
}

@misc{ma2025coevolvingyoufinetuningllm,
      title={Coevolving with the Other You: Fine-Tuning LLM with Sequential Cooperative Multi-Agent Reinforcement Learning}, 
      author={Hao Ma and Tianyi Hu and Zhiqiang Pu and Boyin Liu and Xiaolin Ai and Yanyan Liang and Min Chen},
      year={2025},
      eprint={2410.06101},
      archivePrefix={arXiv},
      primaryClass={cs.AI},
      url={https://arxiv.org/abs/2410.06101}, 
}

@misc{hsieh2024rulerwhatsrealcontext,
      title={RULER: What's the Real Context Size of Your Long-Context Language Models?}, 
      author={Cheng-Ping Hsieh and Simeng Sun and Samuel Kriman and Shantanu Acharya and Dima Rekesh and Fei Jia and Yang Zhang and Boris Ginsburg},
      year={2024},
      eprint={2404.06654},
      archivePrefix={arXiv},
      primaryClass={cs.CL},
      url={https://arxiv.org/abs/2404.06654}, 
}

@misc{shao2024deepseekmathpushinglimitsmathematical,
      title={DeepSeekMath: Pushing the Limits of Mathematical Reasoning in Open Language Models}, 
      author={Zhihong Shao and Peiyi Wang and Qihao Zhu and Runxin Xu and Junxiao Song and Xiao Bi and Haowei Zhang and Mingchuan Zhang and Y. K. Li and Y. Wu and Daya Guo},
      year={2024},
      eprint={2402.03300},
      archivePrefix={arXiv},
      primaryClass={cs.CL},
      url={https://arxiv.org/abs/2402.03300}, 
}

@misc{yao2023treethoughts,
      title={Tree of Thoughts: Deliberate Problem Solving with Large Language Models}, 
      author={Shunyu Yao and Dian Yu and Jeffrey Zhao and Izhak Shafran and Thomas L. Griffiths and Yuan Cao and Karthik Narasimhan},
      year={2023},
      eprint={2305.10601},
      archivePrefix={arXiv},
      primaryClass={cs.CL},
      url={https://arxiv.org/abs/2305.10601}, 
}

@misc{wang2023selfconsistency,
      title={Self-Consistency Improves Chain of Thought Reasoning in Language Models}, 
      author={Xuezhi Wang and Jason Wei and Dale Schuurmans and Quoc Le and Ed Chi and Sharan Narang and Aakanksha Chowdhery and Denny Zhou},
      year={2023},
      eprint={2203.11171},
      archivePrefix={arXiv},
      primaryClass={cs.CL},
      url={https://arxiv.org/abs/2203.11171}, 
}

@misc{liu2023lostmiddle,
      title={Lost in the Middle: How Language Models Use Long Contexts}, 
      author={Nelson F. Liu and Kevin Lin and John Hewitt and Ashwin Paranjape and Michele Bevilacqua and Fabio Petroni and Percy Liang},
      year={2023},
      eprint={2307.03172},
      archivePrefix={arXiv},
      primaryClass={cs.CL},
      url={https://arxiv.org/abs/2307.03172}, 
}

@misc{zhou2020bertlosespatiencefast,
      title={BERT Loses Patience: Fast and Robust Inference with Early Exit}, 
      author={Wangchunshu Zhou and Canwen Xu and Tao Ge and Julian McAuley and Ke Xu and Furu Wei},
      year={2020},
      eprint={2006.04152},
      archivePrefix={arXiv},
      primaryClass={cs.CL},
      url={https://arxiv.org/abs/2006.04152}, 
}

@misc{liu2020fasterdepthadaptivetransformers,
      title={Faster Depth-Adaptive Transformers}, 
      author={Yijin Liu and Fandong Meng and Jie Zhou and Yufeng Chen and Jinan Xu},
      year={2020},
      eprint={2004.13542},
      archivePrefix={arXiv},
      primaryClass={cs.CL},
      url={https://arxiv.org/abs/2004.13542}, 
}

@misc{he2021magicpyramidacceleratinginference,
      title={Magic Pyramid: Accelerating Inference with Early Exiting and Token Pruning}, 
      author={Xuanli He and Iman Keivanloo and Yi Xu and Xiang He and Belinda Zeng and Santosh Rajagopalan and Trishul Chilimbi},
      year={2021},
      eprint={2111.00230},
      archivePrefix={arXiv},
      primaryClass={cs.CL},
      url={https://arxiv.org/abs/2111.00230}, 
}

@misc{raposo2024mixtureofdepthsdynamicallyallocatingcompute,
      title={Mixture-of-Depths: Dynamically allocating compute in transformer-based language models}, 
      author={David Raposo and Sam Ritter and Blake Richards and Timothy Lillicrap and Peter Conway Humphreys and Adam Santoro},
      year={2024},
      eprint={2404.02258},
      archivePrefix={arXiv},
      primaryClass={cs.LG},
      url={https://arxiv.org/abs/2404.02258}, 
}

@misc{yao2022react,
      title={ReAct: Synergizing Reasoning and Acting in Language Models}, 
      author={Shunyu Yao and Jeffrey Zhao and Dian Yu and Nan Du and Izhak Shafran and Karthik Narasimhan and Yuan Cao},
      year={2023},
      eprint={2210.03629},
      archivePrefix={arXiv},
      primaryClass={cs.CL},
      url={https://arxiv.org/abs/2210.03629}, 
}

@inproceedings{NIPS2003_c8067ad1,
 author = {Chang, Yu-han and Ho, Tracey and Kaelbling, Leslie},
 booktitle = {Advances in Neural Information Processing Systems},
 editor = {S. Thrun and L. Saul and B. Sch\"{o}lkopf},
 pages = {},
 publisher = {MIT Press},
 title = {All learning is Local: Multi-agent Learning in Global Reward Games},
 url = {https://proceedings.neurips.cc/paper_files/paper/2003/file/c8067ad1937f728f51288b3eb986afaa-Paper.pdf},
 volume = {16},
 year = {2003}
}

@misc{foerster2024counterfactualmultiagentpolicygradients,
      title={Counterfactual Multi-Agent Policy Gradients}, 
      author={Jakob Foerster and Gregory Farquhar and Triantafyllos Afouras and Nantas Nardelli and Shimon Whiteson},
      year={2024},
      eprint={1705.08926},
      archivePrefix={arXiv},
      primaryClass={cs.AI},
      url={https://arxiv.org/abs/1705.08926}, 
}

@misc{sunehag2017valuedecompositionnetworkscooperativemultiagent,
      title={Value-Decomposition Networks For Cooperative Multi-Agent Learning}, 
      author={Peter Sunehag and Guy Lever and Audrunas Gruslys and Wojciech Marian Czarnecki and Vinicius Zambaldi and Max Jaderberg and Marc Lanctot and Nicolas Sonnerat and Joel Z. Leibo and Karl Tuyls and Thore Graepel},
      year={2017},
      eprint={1706.05296},
      archivePrefix={arXiv},
      primaryClass={cs.AI},
      url={https://arxiv.org/abs/1706.05296}, 
}

@misc{rashid2018qmixmonotonicvaluefunction,
      title={QMIX: Monotonic Value Function Factorisation for Deep Multi-Agent Reinforcement Learning}, 
      author={Tabish Rashid and Mikayel Samvelyan and Christian Schroeder de Witt and Gregory Farquhar and Jakob Foerster and Shimon Whiteson},
      year={2018},
      eprint={1803.11485},
      archivePrefix={arXiv},
      primaryClass={cs.LG},
      url={https://arxiv.org/abs/1803.11485}, 
}

@inproceedings{NEURIPS2020_1f89885d,
 author = {Stiennon, Nisan and Ouyang, Long and Wu, Jeffrey and Ziegler, Daniel and Lowe, Ryan and Voss, Chelsea and Radford, Alec and Amodei, Dario and Christiano, Paul F},
 booktitle = {Advances in Neural Information Processing Systems},
 editor = {H. Larochelle and M. Ranzato and R. Hadsell and M.F. Balcan and H. Lin},
 pages = {3008--3021},
 publisher = {Curran Associates, Inc.},
 title = {Learning to summarize with human feedback},
 url = {https://proceedings.neurips.cc/paper_files/paper/2020/file/1f89885d556929e98d3ef9b86448f951-Paper.pdf},
 volume = {33},
 year = {2020}
}

@misc{schulman2017trustregionpolicyoptimization,
      title={Trust Region Policy Optimization}, 
      author={John Schulman and Sergey Levine and Philipp Moritz and Michael I. Jordan and Pieter Abbeel},
      year={2017},
      eprint={1502.05477},
      archivePrefix={arXiv},
      primaryClass={cs.LG},
      url={https://arxiv.org/abs/1502.05477}, 
}

@misc{schulman2017proximalpolicyoptimizationalgorithms,
      title={Proximal Policy Optimization Algorithms}, 
      author={John Schulman and Filip Wolski and Prafulla Dhariwal and Alec Radford and Oleg Klimov},
      year={2017},
      eprint={1707.06347},
      archivePrefix={arXiv},
      primaryClass={cs.LG},
      url={https://arxiv.org/abs/1707.06347}, 
}

@misc{hendrycks2021measuringmathematicalproblemsolving,
      title={Measuring Mathematical Problem Solving With the MATH Dataset}, 
      author={Dan Hendrycks and Collin Burns and Saurav Kadavath and Akul Arora and Steven Basart and Eric Tang and Dawn Song and Jacob Steinhardt},
      year={2021},
      eprint={2103.03874},
      archivePrefix={arXiv},
      primaryClass={cs.LG},
      url={https://arxiv.org/abs/2103.03874}, 
}

@misc{ho2020constructingmultihopqadataset,
      title={Constructing A Multi-hop QA Dataset for Comprehensive Evaluation of Reasoning Steps}, 
      author={Xanh Ho and Anh-Khoa Duong Nguyen and Saku Sugawara and Akiko Aizawa},
      year={2020},
      eprint={2011.01060},
      archivePrefix={arXiv},
      primaryClass={cs.CL},
      url={https://arxiv.org/abs/2011.01060}, 
}

@misc{trivedi2022musiquemultihopquestionssinglehop,
      title={MuSiQue: Multihop Questions via Single-hop Question Composition}, 
      author={Harsh Trivedi and Niranjan Balasubramanian and Tushar Khot and Ashish Sabharwal},
      year={2022},
      eprint={2108.00573},
      archivePrefix={arXiv},
      primaryClass={cs.CL},
      url={https://arxiv.org/abs/2108.00573}, 
}

@misc{yuan2023largelanguagemodelsperform,
      title={How well do Large Language Models perform in Arithmetic tasks?}, 
      author={Zheng Yuan and Hongyi Yuan and Chuanqi Tan and Wei Wang and Songfang Huang},
      year={2023},
      eprint={2304.02015},
      archivePrefix={arXiv},
      primaryClass={cs.CL},
      url={https://arxiv.org/abs/2304.02015}, 
}

@misc{zheng2023judging,
      title={Judging LLM-as-a-Judge with MT-Bench and Chatbot Arena}, 
      author={Lianmin Zheng and Wei-Lin Chiang and Ying Sheng and Siyuan Zhuang and Zhanghao Wu and Yonghao Zhuang and Zi Lin and Zhuohan Li and Dacheng Li and Eric P. Xing and Hao Zhang and Joseph E. Gonzalez and Ion Stoica},
      year={2023},
      eprint={2306.05685},
      archivePrefix={arXiv},
      primaryClass={cs.CL},
      url={https://arxiv.org/abs/2306.05685}, 
}

@InProceedings{pmlr-v162-li22w,
  title = 	 {Difference Advantage Estimation for Multi-Agent Policy Gradients},
  author =       {Li, Yueheng and Xie, Guangming and Lu, Zongqing},
  booktitle = 	 {Proceedings of the 39th International Conference on Machine Learning},
  pages = 	 {13066--13085},
  year = 	 {2022},
  editor = 	 {Chaudhuri, Kamalika and Jegelka, Stefanie and Song, Le and Szepesvari, Csaba and Niu, Gang and Sabato, Sivan},
  volume = 	 {162},
  series = 	 {Proceedings of Machine Learning Research},
  month = 	 {17--23 Jul},
  publisher =    {PMLR},
  url = 	 {https://proceedings.mlr.press/v162/li22w.html}
}

@misc{wang2024mathshepherdverifyreinforcellms,
      title={Math-Shepherd: Verify and Reinforce LLMs Step-by-step without Human Annotations}, 
      author={Peiyi Wang and Lei Li and Zhihong Shao and R. X. Xu and Damai Dai and Yifei Li and Deli Chen and Y. Wu and Zhifang Sui},
      year={2024},
      eprint={2312.08935},
      archivePrefix={arXiv},
      primaryClass={cs.AI},
      url={https://arxiv.org/abs/2312.08935}, 
}

@article{sheng2024hybridflow,
   title={HybridFlow: A Flexible and Efficient RLHF Framework},
   url={http://dx.doi.org/10.1145/3689031.3696075},
   DOI={10.1145/3689031.3696075},
   booktitle={Proceedings of the Twentieth European Conference on Computer Systems},
   publisher={ACM},
   author={Sheng, Guangming and Zhang, Chi and Ye, Zilingfeng and Wu, Xibin and Zhang, Wang and Zhang, Ru and Peng, Yanghua and Lin, Haibin and Wu, Chuan},
   year={2025},
   month=mar, pages={1279–1297},
   collection={EuroSys ’25}
}

@inproceedings{kwon2023efficient,
      title={Efficient Memory Management for Large Language Model Serving with PagedAttention}, 
      author={Woosuk Kwon and Zhuohan Li and Siyuan Zhuang and Ying Sheng and Lianmin Zheng and Cody Hao Yu and Joseph E. Gonzalez and Hao Zhang and Ion Stoica},
      year={2023},
      eprint={2309.06180},
      archivePrefix={arXiv},
      primaryClass={cs.LG},
      url={https://arxiv.org/abs/2309.06180}, 
}

@misc{leviathan2023speculativedecoding,
      title={Fast Inference from Transformers via Speculative Decoding}, 
      author={Yaniv Leviathan and Matan Kalman and Yossi Matias},
      year={2023},
      eprint={2211.17192},
      archivePrefix={arXiv},
      primaryClass={cs.LG},
      url={https://arxiv.org/abs/2211.17192}, 
}

@inproceedings{Zhang_2024,
   title={Draft\& Verify: Lossless Large Language Model Acceleration via Self-Speculative Decoding},
   url={http://dx.doi.org/10.18653/v1/2024.acl-long.607},
   DOI={10.18653/v1/2024.acl-long.607},
   booktitle={Proceedings of the 62nd Annual Meeting of the Association for Computational Linguistics (Volume 1: Long Papers)},
   publisher={Association for Computational Linguistics},
   author={Zhang, Jun and Wang, Jue and Li, Huan and Shou, Lidan and Chen, Ke and Chen, Gang and Mehrotra, Sharad},
   year={2024},
   pages={11263–11282} }

@misc{cai2024medusasimplellminference,
      title={Medusa: Simple LLM Inference Acceleration Framework with Multiple Decoding Heads}, 
      author={Tianle Cai and Yuhong Li and Zhengyang Geng and Hongwu Peng and Jason D. Lee and Deming Chen and Tri Dao},
      year={2024},
      eprint={2401.10774},
      archivePrefix={arXiv},
      primaryClass={cs.LG},
      url={https://arxiv.org/abs/2401.10774}, 
}

@article{zhou2024archer,
      title={ArCHer: Training Language Model Agents via Hierarchical Multi-Turn RL}, 
      author={Yifei Zhou and Andrea Zanette and Jiayi Pan and Sergey Levine and Aviral Kumar},
      year={2024},
      eprint={2402.19446},
      archivePrefix={arXiv},
      primaryClass={cs.LG},
      url={https://arxiv.org/abs/2402.19446}, 
}

@article{qi2024webrl,
      title={WebRL: Training LLM Web Agents via Self-Evolving Online Curriculum Reinforcement Learning}, 
      author={Zehan Qi and Xiao Liu and Iat Long Iong and Hanyu Lai and Xueqiao Sun and Wenyi Zhao and Yu Yang and Xinyue Yang and Jiadai Sun and Shuntian Yao and Tianjie Zhang and Wei Xu and Jie Tang and Yuxiao Dong},
      year={2025},
      eprint={2411.02337},
      archivePrefix={arXiv},
      primaryClass={cs.CL},
      url={https://arxiv.org/abs/2411.02337}, 
}

@article{yuan2025reinforce,
      title={Reinforce LLM Reasoning through Multi-Agent Reflection}, 
      author={Yurun Yuan and Tengyang Xie},
      year={2025},
      eprint={2506.08379},
      archivePrefix={arXiv},
      primaryClass={cs.LG},
      url={https://arxiv.org/abs/2506.08379}, 
}

@article{shinn2023reflexion,
      title={Reflexion: Language Agents with Verbal Reinforcement Learning}, 
      author={Noah Shinn and Federico Cassano and Edward Berman and Ashwin Gopinath and Karthik Narasimhan and Shunyu Yao},
      year={2023},
      eprint={2303.11366},
      archivePrefix={arXiv},
      primaryClass={cs.AI},
      url={https://arxiv.org/abs/2303.11366}, 
}

@article{wang2023voyager,
      title={Voyager: An Open-Ended Embodied Agent with Large Language Models}, 
      author={Guanzhi Wang and Yuqi Xie and Yunfan Jiang and Ajay Mandlekar and Chaowei Xiao and Yuke Zhu and Linxi Fan and Anima Anandkumar},
      year={2023},
      eprint={2305.16291},
      archivePrefix={arXiv},
      primaryClass={cs.AI},
      url={https://arxiv.org/abs/2305.16291}, 
}

@article{schick2023toolformer,
      title={Toolformer: Language Models Can Teach Themselves to Use Tools}, 
      author={Timo Schick and Jane Dwivedi-Yu and Roberto Dessì and Roberta Raileanu and Maria Lomeli and Luke Zettlemoyer and Nicola Cancedda and Thomas Scialom},
      year={2023},
      eprint={2302.04761},
      archivePrefix={arXiv},
      primaryClass={cs.CL},
      url={https://arxiv.org/abs/2302.04761}, 
}

@dataset{aime_1983_2024,
  author = {Hemish Veeraboina},
  title = {AIME Problem Set 1983-2024},
  year = {2023},
  publisher = {Kaggle},
  url = {https://www.kaggle.com/datasets/hemishveeraboina/aime-problem-set-1983-2024}
}

@misc{singh2025agenticretrievalaugmentedgenerationsurvey,
      title={Agentic Retrieval-Augmented Generation: A Survey on Agentic RAG}, 
      author={Aditi Singh and Abul Ehtesham and Saket Kumar and Tala Talaei Khoei},
      year={2025},
      eprint={2501.09136},
      archivePrefix={arXiv},
      primaryClass={cs.AI},
      url={https://arxiv.org/abs/2501.09136}, 
}
\bibliographystyle{icml2026}




\end{document}